\documentclass[10pt,twocolumn,letterpaper]{article}

\usepackage{3dv}
\usepackage{times}
\usepackage{epsfig}
\usepackage{graphicx}
\usepackage{amsmath}
\usepackage{amssymb}
\usepackage{multirow}
\usepackage{dsfont}
\usepackage{dblfloatfix}
\usepackage{arydshln}
\usepackage[pagebackref=true,breaklinks=true,letterpaper=true,colorlinks,bookmarks=false]{hyperref}

\newcommand{\head}[1]{\noindent\textbf{#1}}
\newcommand\blfootnote[1]{%
  \begingroup
  \renewcommand\thefootnote{}\footnote{#1}%
  \addtocounter{footnote}{-1}%
  \endgroup
}

\threedvfinalcopy 

\def\threedvPaperID{****} 

\ifthreedvfinal\fi
\begin{document}

\title{Reconstructing Action-Conditioned Human-Object Interactions \\Using Commonsense Knowledge Priors}

\author{Xi Wang\textsuperscript{1}* \quad Gen Li\textsuperscript{1}* \quad Yen-Ling Kuo\textsuperscript{2}  \quad Muhammed Kocabas\textsuperscript{1,3} \quad Emre Aksan\textsuperscript{1} \quad Otmar Hilliges\textsuperscript{1}
\\ 
    \normalsize\textsuperscript{1}ETH Zürich \quad 
    \textsuperscript{2}Massachusetts Institute of Technology \quad
    \normalsize\textsuperscript{3}Max Planck Institute for Intelligent Systems \\
}

\maketitle
\thispagestyle{empty}

\begin{abstract}
We present a method for inferring diverse 3D models of human-object interactions from images.
Reasoning about how humans interact with objects in complex scenes from a single 2D image is a challenging task given ambiguities arising from the loss of information through projection. 
In addition, modeling 3D interactions requires the generalization ability towards diverse object categories and interaction types. 
We propose an action-conditioned modeling of interactions that allows us to infer diverse 3D arrangements of humans and objects without supervision on contact regions or 3D scene geometry. 
Our method extracts high-level commonsense knowledge from large language models (such as GPT-3), and applies them to perform 3D reasoning of human-object interactions. 
Our key insight is priors extracted from large language models can help in reasoning about human-object contacts from textural prompts only.
We quantitatively evaluate the inferred 3D models on a large human-object interaction dataset and show how our method leads to better 3D reconstructions. We further qualitatively evaluate the effectiveness of our method on real images and demonstrate its generalizability towards interaction types and object categories. 
\blfootnote{*Denotes equal contribution}
\end{abstract}

\section{Introduction}

Humans interact with objects in diverse ways. Take a chair for example. Someone could sit on a chair to read a book, another could rest their hand on a chair, while others might stand on a chair to reach a high shelf. The ability to understand diverse interactions between humans and objects is of critical importance for practical applications that require understanding human actions, such as content creation in VR/AR, 3D scene understanding and personalized robotics. Recent works~\cite{weng2021holistic, xie2022chore, zhang2020perceiving} have demonstrated impressive results in learning human-object interactions from images. Such tasks require not only an accurate reconstruction of body shape and pose, object shape and pose, but also joint reasoning about contacts between human bodies and objects. 

A key challenge is how to efficiently infer the diverse interactions between humans and objects. As in the previous example, one can sit, lean on, or stand on a chair. We need to distinguish between these different interactions in order to successfully reconstruct 3D models of humans and objects. Current publicly available approaches do not explicitly model interaction diversity, despite its importance. Pure image-based learning approaches~\cite{zhang2020perceiving} directly reconstruct 3D interaction models from 2D content by incorporating holistically defined contact rules and priors on object size. These rules provide contact regions at part levels of human bodies and objects. More precisely, they tell which part of the body is in contact with which part of the object. However, the establishment of these contact rules requires manual annotations on pairs of bodies and objects, which poses a fundamental limitation on their scalability~\cite{xie2022chore, zhang2022couch}. Instead, most methods resort to learning from 3D data~\cite{bhatnagar2022behave, hassan2019resolving, savva2016pigraphs, starke2019neural}. While recordings of humans interacting with objects may provide increased diversity, collecting a large-scale dataset demands tremendous effort and often requires specialized hardware (e.g. MoCap, RGBD cameras). Furthermore, the number of object categories and the diversity of interactions are often limited by practical constraints. 

We present an optimization approach that infers diverse human-object interactions through an action-conditioned formulation. Our key idea is to use knowledge extracted from large language models (LLMs) in low-level computer vision tasks. Specifically, our method infers the underlying action type directly from human poses, allowing us to distinguish between sitting and standing on a chair. This action condition is further used in the query of LLMs, where we obtain part-wise contact information between humans and objects, waiving the requirement of manual annotations. Following Phosa~\cite{zhang2020perceiving}, we aim to directly infer from images, as they provide a rich source of information of natural, diverse and inclusive types of interactions. We propose a two-stage approach which first estimates shapes and poses of humans and objects independently, and jointly reasons about human-object interactions and their 3D spatial arrangements in the second optimization step. 

Reconstructing 3D humans and objects from images is, however, extremely challenging. Apart from the large degree of scene clutter, humans and objects frequently occlude each other during interaction. We observe that the prediction of body shapes and poses is significantly more robust than object shape and pose estimation, as object shapes and poses have a considerably large variation. This is possible in practice thanks to the large effort on human body and pose modeling~\cite{Kocabas_PARE_2021, kocabas2021spec, SMPL:2015, SMPL-X:2019, MANO:SIGGRAPHASIA:2017}. We thus use body poses to represent the underlying actions under the assumption that the variation of poses within each interaction type is smaller than the variation of poses across different interaction types. 

Our experiments demonstrate the value of using knowledge extracted from LLMs for action-conditioned interaction modeling. Furthermore, a user study performed on the generated labels shows that the priors derived from LLMs are mostly aligned with humans' priors. Finally, we show that our proposed approach can accurately infer 3D humans and objects from images under diverse interactions using both quantitative and qualitative evaluations.   
To summarize, our contributions are:

\begin{itemize}
    \item An optimization-based framework that reconstructs 3D human-object interaction models using an action-conditioned formulation which significantly expands the scope and diversity of interactions that can be inferred from images,
    \item To avoid costly human annotations, we show that commonsense knowledge extracted from LLMs can provide useful priors for reasoning about human-object contacts from textural prompts only,
    \item Finally, we show that commonsense knowledge bases can be used to define an action space that effectively covers different types of interactions, and a retrieval-based action recognition module allows us to estimate the underlying actions by comparing body pose similarities.
\end{itemize}

\section{Related Work}
\head{3D human and object reconstruction}

Accurate human-object contact estimation plays a crucial role in interaction modeling. 
Previous approaches have explicitly used contact regions between human bodies and objects to reason about their spatial arrangements in 3D. 
Prox~\cite{hassan2019resolving} estimates 3D human pose from a single RGB image with known 3D scene information. It encourages contacts between human bodies and objects that are in close proximity and  formulates it as a heuristic contact constraint. 
Similarly, Phosa~\cite{zhang2020perceiving} infers shapes and 3D spatial arrangements of human bodies and objects from images, but without access to 3D scene information. A part-based interaction loss is used to pull pairs of object-specific contact regions together, which are obtained through manual annotations.
HolisticMesh~\cite{weng2021holistic} introduces a learning model that jointly reconstructs 3D human bodies and objects from RGB images. A set of holistic objectives are imposed to ensure global consistency, including estimating camera pose and room layout. 
Recent work Mover~\cite{yi2022human} shows how to improve 3D scene reconstruction from RGB videos by leveraging human scene contact constraints. Similar to HolisticMesh, Mover includes the optimization over camera pose and ground-plane pose, and uses a contact loss that considers object normal directions.  
While these methods are powerful, they require manually annotated contact points either for individual objects~\cite{yi2022human, zhang2020perceiving}, or a single body annotation is used for all types of contact~\cite{hassan2019resolving, weng2021holistic}. Thus it is difficult to generalize to large set of objects. One contribution of our work is to use LLMs to identify contact regions between human bodies and objects.
   
Several recent work specifically aims to overcome the limitations imposed by human annotations.  
The newly released BEHAVE~\cite{bhatnagar2022behave} dataset is specifically designed for this purpose. It provides a large dataset with multi-view RGBD frames together with annotated contacts between bodies and objects. 
Chore~\cite{xie2022chore} uses the BEHAVE dataset to learn a neural reconstruction of human and object from a single RGB image. Both human and object are represented by unsigned distance fields and the model is trained to predict contact points from input images by using priors learned from data.  
It is not clear that how well these methods perform on objects that are not seen in the dataset. This fundamentally limits their ability to generalize towards new object categories, another point our work addresses.  

On the other hand, humans interact with objects in various diverse way. 
Couch~\cite{zhang2022couch} studies the problem of synthesizing diverse human-object interactions. Specifically, it focuses on the interaction between humans and chairs, and train a model to generate plausible motion sequences with or without predefined contact points on the object. In contrast, we use 
an interaction space defined by plausible actions to learn action-conditioned, diverse human-object interactions.

\begin{figure*}[t]
\begin{center}
   \includegraphics[width=0.95\textwidth]{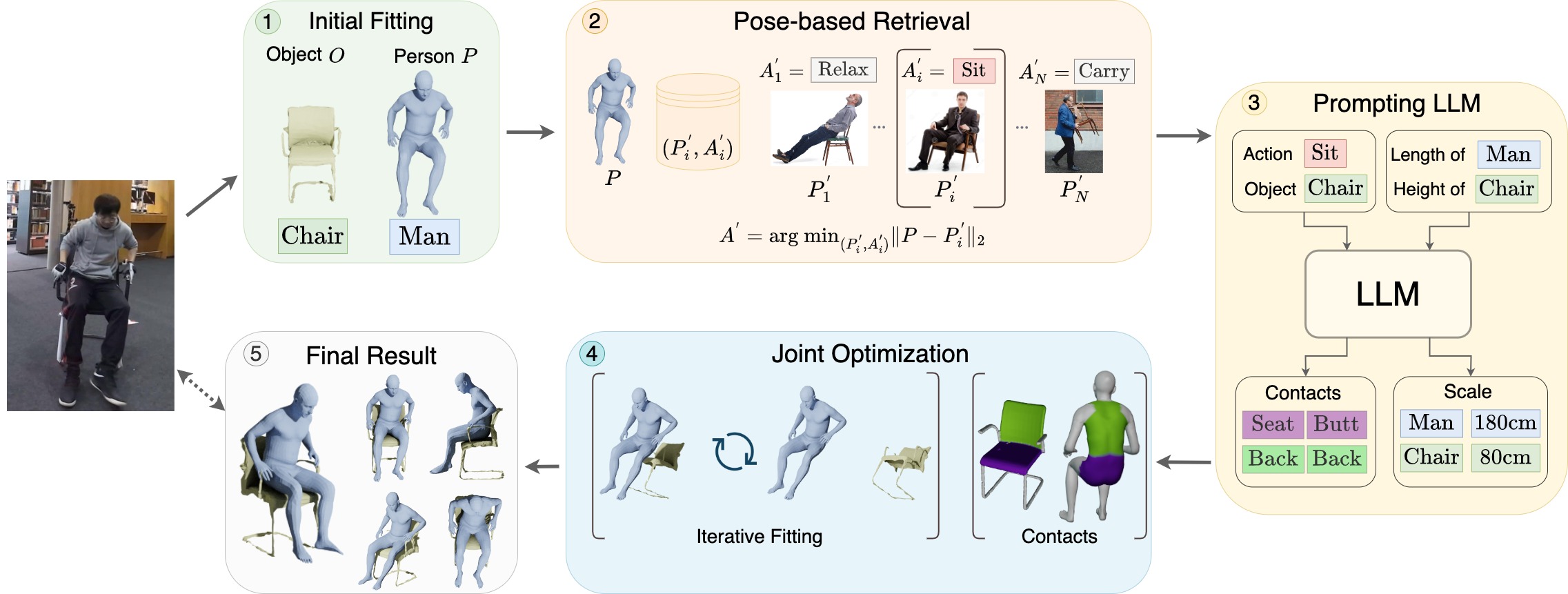}
\end{center}
\caption{\textbf{Overview of the approach.} Our method first independently estimates pose and shape of human body and object (1). The estimated body pose is then used to retrieve a closest pose from a database alongside a known action type (2). The retrieved action label is then used to prompt LLMs for action-conditioned contact information and object size (3). A final optimization jointly reasons about spatial arrangements and shapes of 3D humans and objects (4). }
\label{fig:overview}
\end{figure*}

\head{Language guided learning}
Humans use language to flexibly refer to objects, relations, and events in our world.
Prior experiments suggests that language influences many aspects of human cognition including spatial reasoning~\cite{li2002turning} and how visual concepts can be acquired~\cite{jones1991object}.
Taking this intuition, many machine learning domains have explored ways to use language to guide the learning process.

Language has been combined with different modalities to improve the learned representations or task performance.
In computer vision, image captions are used to learn the representations of images~\cite{radford2021learning,rao2021denseclip} and to reason about the visual relationships~\cite{liao2019natural}
Language is also used as an additional input to guide tasks such as video summarization~\cite{narasimhan2021clip}.
In robotics or policy learning, the agents not only follow instructions, but also learn to update semantic map for robot manipulation~\cite{patki2020language}, trajectory reshaping~\cite{bucker2022reshaping}, and new skills with language inputs~\cite{sharma2022skill}.

In addition to take language as an input modality, language can provide information about the structure of tasks or the knowledge for solving a problem.
Recent research has designed the filters or the structure of neural module networks guided by language to enable generalization to novel combinations of known concepts~\cite{akula2021robust,kuo:compositional-gscan2021}.
LLMs have been shown to encode significant amount of knowledge~\cite{petroni-etal-2019-language} that is useful to guide several tasks, e.g. procedural knowledge for planning actions~\cite{huang2022language}.
It is also possible to combine LLMs with other pretrained models for zero-shot multimodal reasoning~\cite{zeng2022socratic}.

Our approach similarly considers language models as the source of information to guide the 3D reconstruction.
We show that the priors derived from LLMs, e.g. object sizes and human-object contact regions, 
are useful heuristics to guide the joint optimization of human and body poses.
The use of LLMs as reconstruction priors further enables generalization to the actions and objects that have never been considered before as it is hard to collect or annotate such knowledge for any type of human-object interactions.

\section{Method Overview}
\label{sec:method}

We aim to infer 3D models of human-object interactions from 2D images, as capturing 3D human-object interactions is difficult and there are few diverse 3D interaction data, outside of the very recent BEHAVE dataset~\cite{bhatnagar2022behave}.  
In this paper, we define this 3D model to be spatial arrangements and shapes of humans and objects.
Since the possible human-object interactions is constrained by human actions, we perform inference by conditioning the 3D reconstruction on the recognized action type.

Figure~\ref{fig:overview} shows an overview of our approach.
Given an input image, we obtain initial estimates for humans and objects separately (in Sec.~\ref{sec:poseinit}).
Next, the obtained 3D human poses are used to retrieve the corresponding action types (in Sec.~\ref{sec:templatematching})
We then use the recognized action types to query LLMs for the part-level contact information (in Sec.~\ref{sec:llmquery}), which is used as prior to infer 3D human-object interactions.
Lastly, 3D human and object pose parameters are jointly optimized through reasoning about their spatial arrangements using the contact information derived from LLMs (in Sec.~\ref{sec:optimization}).  

Our framework is related to the two-stage approach of Phosa~\cite{zhang2020perceiving} which first performs an independent reconstruction of humans and objects, and reasons about their contacts and 3D spatial arrangements in the second joint optimization step.
Unlike Phosa, we use the recognized action types and the knowledge extracted from LLMs to guide the optimization.
These components greatly help inference of diverse types of interactions and enable generalization to a wide range of objects and actions. 
In contrast, prior approaches require a prior knowledge of a small set of objects.

\subsection{Independent pose and shape estimation}
\label{sec:poseinit}

\head{Estimating human body pose and shape}
To obtain initial estimates for human pose and shape from an input image, we use SMPLify~\cite{SMPL-X:2019}, an optimization-based fitting method, initialized from a state-of-the-art human pose and shape regression method PARE~\cite{Kocabas_PARE_2021}. Given an input image, PARE regresses the pose ($\theta$) and shape ($\beta$) parameters of the parametric body model SMPL. These estimated parameters are then used to initialize the SMPLify optimization. SMPLify uses the objective function in Eq.~\ref{eq:smplify} to fit the SMPL body model to 2D keypoints estimated using an off-the-shelf 2D keypoint estimation model~\cite{mmpose2020}.
\begin{equation}
\label{eq:smplify}
    E(\beta, \theta) = E_{J} + E_{\theta} + E_{\beta},
\end{equation}
where $\beta, \theta$ are SMPL shape and pose parameters, $E_{\theta}$ and $E_\beta$ are pose and shape prior terms. $E_{J}$ is the data term that measures the difference between the detected $\mathcal{J}_{\mathit{2D}}$ and the 2D projection of the estimated $\hat{\mathcal{J}}_{\mathit{3D}}$ joints under perspective projection $\Pi$:
\begin{equation}
\label{eq:persp-proj}
    E_{J} = \Big|\Big| \Pi (\hat{\mathcal{J}}_{\mathit{3D}}) - \mathcal{J}_{\mathit{2D}} \Big|\Big|^{2}_{2}.
\end{equation}

\head{Estimating object pose and shape}
Estimating object pose and shape from real images has been shown to be a difficult task~\cite{henzler2019escaping, niemeyer2020differentiable, richter2018matryoshka}, in particular due to the large variations of shape in man-made objects. We find this to be the case in learning 3D human-object interactions as well. 
Moreover, the task gets more challenging when objects are off-center, partially occluded, or if multiple objects exist in one image. Accordingly, we reformulate the task of object shape estimation as shape matching. We consider several representative shape exemplars for each object category and select the top-$5$ models that best match the corresponding 2D image. 

We begin by detecting objects in an image using the PointRend object detector~\cite{kirillov2020pointrend}, and obtain bounding boxes, segmentation masks and semantic labels for each object. A differentiable renderer~\cite{kato2018neural} is then used to optimize for 6-DoF object poses. Note that an intrinsic scale for each object category is required to remove depth ambiguity, and we obtain such a scale by prompting LLMs (see details in Sec.~\ref{sec:llmquery}).

\subsection{Retrieval-based action recognition}
\label{sec:templatematching}
The actions humans take influence the underlying human-object interactions.
So far, we have reconstructed 3D humans and objects. However, to correctly infer their arrangements, we will extract the corresponding action type of an interaction in an image.
We hypothesize that while interacting with objects, the associated actions are captured by the corresponding body poses.
Instead of directly using an action recognition method, our method extracts actions by selecting an exemplar pose from a set of $N$ body poses with known action types for each detected body pose in the input images.
This choice is motivated by our goal to model diverse interactions.
The set of $N$ body poses denote the potential interactions for each human-object pair.

In summary, our method uses body poses to index the interaction space and identify the associated actions. 
Given an input image, we first reconstruct body shapes and poses for all detected human instances. 
The obtained body poses are then used to access a database via a nearest neighbor lookup and learn the associated action types based on retrieved interaction models. 
More specifically, given a body pose $p$, we choose the most similar pose $p'$ from a set of body poses whose action labels are known. Here we select a single pose instead of top-$k$. We then transfer the associated action label of $p'$ to $p$. Since the parametric SMPL model has disentangled body shape and pose parameters, we can use the pose parameters $\theta$ to directly compare different body poses. More precisely, we use the Euclidean distance between the corresponding joint rotations as the similarity metric and build a k-d tree to search for the closest pose.
By combining this retrieval-based action recognition strategy and the reconstruction pipeline, we obtain a framework that can reconstruct diverse interaction models for individual object categories. 

\head{Building the interaction database}
To improve the reconstruction quality, we propose to complement the optimization pipeline with a retrieval-based approach and use the interaction database as an explicit memory.
We define the possible actions humans can interact with an object using ConceptNet~\cite{speer2017conceptnet}, a large crowd-sourced commonsense knowledge base.
We consider actions that connect to the object using ``UsedFor'' or ``RelatedTerms'' relationships and have weight greater or equal to 1 to denote the most relevant actions.
On average, we have 14 actions for each object.
For each action obtained from ConceptNet, we construct simple sentences by combining it with the related object noun, e.g., ``woman sit chair'', and use them to search and download images. We filter out candidates based on its size and content. More precisely, we keep images that have more than $300$ pixels in both dimensions and contain exactly one person and one object.
We also discard images that contain invalid human-object interactions by computing the intersection over union (IoU) of their 2D bounding boxes. These filters discard more than $80\%$ of the candidates.   
Most of the remaining images have both humans and objects that are clearly visible at the center.

\subsection{Deriving interaction priors from LLMs}
\label{sec:llmquery}
To learn diverse human-object interactions, it requires an understanding of knowledge such as the size of an object and how an object can be interacted with.
The knowledge about the possible types of interactions largely influences how we formulate the optimization of 3D human-object interaction (in Sec.~\ref{sec:optimization}).
While there are multiple ways humans can interact with an object, knowing the current action largely helps in distinguishing various interactions.
For example, humans may touch different part of chairs for ``sitting on a chair'' versus ``standing on a chair''.
We show that such knowledge is encoded in LLMs and can be accessed through probing via fill-in-the-blank (cloze) prompts~\cite{taylor1953cloze}.
We design a set of prompts (see the Supplementary Material 
for the templates) to derive the following interaction priors that are used in optimization:
\begin{itemize}
    \item \textbf{Size of an object}: Given the name of an object, the LLM generates the estimated size of that object in meters. This prior allows us to scale the object relative to human bodies correctly.
    \item \textbf{Possible contact regions of body and object parts}: Given the object and the action to interact with the object, we first prompt the LLM to generate the list of possible contacts between a human body and the object. For example, our system takes ``stand on a chair'' as input to generate ``foot and chair seat'' as a possible contact. We then use a second prompt to select how the generated body and object parts are mapped to the labels used in PartNet or the SMPL body model. Pairs of contact parts are stored in the interaction map $\mathcal{I}$.
\end{itemize}
Note that Phosa~\cite{zhang2020perceiving} has shown the importance of having a reasonable object size for the reconstruction task and a good initialization helps to reason about contact and resolve depth ambiguity. 

\subsection{Joint optimization}
\label{sec:optimization}

Separated reconstruction of human bodies and objects often fails to produce satisfying results. As demonstrated in previous methods~\cite{hassan2019resolving, weng2021holistic, yi2022human, zhang2020perceiving}, people may end up sitting in the air or intersecting with objects. Thus,we formulate joint optimization to reason about contacts and spatial arrangements in 3D. We optimize for rigid transformations (isotropic scaling, translation, and rotation) to be applied to 3D objects and allow a simple scaling variable for 3D humans. The following objective loss functions are optimized. 

We consider a reprojection loss $\mathcal{L}_{reprojection}$ that fits the projected objects in 2D to the detected object masks while preserving the boundaries~\cite{zhang2020perceiving}. 
The scale loss $\mathcal{L}_{scale}$ computes the squared distance $\left\Vert s'_c - \bar{s}_c \right\Vert$ between the current scale $s'_c$ and the scale $\bar{s}_c$ obtained through LLMs. The scale can be of either object or human $c\in[\mathcal{O}, \mathcal{H}]$. 

Given the part-level contacts obtained from LLMs, we formulate a contact loss $\mathcal{L}_{contact}$ for each pair of interacting parts of human and object. Intuitively, the contact loss attempts to bring closer parts of humans and objects that interact. 
\begin{equation}
    \mathcal{L}_{contact} = \sum_{(i,j)\in\mathcal{I}}\mathds{1}(\mathbf{n}_{\mathcal{P}_{h}^i}, \mathbf{n}_{\mathcal{P}_{o}^j}) d_{CD}(\mathcal{P}_{h}^i, \mathcal{P}_{o}^j),
\end{equation}
where $\mathcal{P}_{h}$ and $\mathcal{P}_{o}$ are parts of human and object, respectively. $(i, j) \in \mathcal{I}$ is a pair of contact parts from the interaction map $\mathcal{I}$. The indicator function 
$\mathds{1}(\mathbf{n}_{\mathcal{P}_{h}^i}, \mathbf{n}_{\mathcal{P}_{o}^j})$ identifies whether the two parts have valid contacts or not. $d_{CD}$ is the one-way Chamfer loss~\cite{gavrila2000pedestrian} that computes the distance of each vertex in object part to the nearest vertex in the body part. 

In addition, we use constraints on surface normals to define meaningful contact surfaces. We assume that only surfaces that have normals pointing in opposite direction can have valid contacts. This assumption is formulated as a normal loss $\mathcal{L}_{normal}$, 
\begin{equation}
\mathcal{L}_{normal} = \sum_{(i,j)\in\mathcal{I}}\mathds{1}(\mathbf{n}_{\mathcal{P}_{h}^i}, \mathbf{n}_{\mathcal{P}_{o}^j})(1 + d_{\cos}(\mathbf{n}_{\mathcal{P}_{h}^i}, \mathbf{n}_{\mathcal{P}_{o}^j})).
\end{equation}
where $d_{\cos}(\mathbf{a}, \mathbf{b})=\frac{\mathbf{a}\cdot \mathbf{b}}{|\mathbf{a}||\mathbf{b}|}$ is the cosine similarity between two normal vectors.

To avoid interpenetration between humans and objects, we add a penetration loss $\mathcal{L}_{penetration}$. We compute $f$, a sign distance field (SDF), for each human body, and discretize the space by $8^3$ grid cells. $\mathcal{L}_{penetration}$ penalizes all object parts that are inside of the body, denoted by $f(\mathbf{v}_k)<0$. $f$ is defined negative at points inside of the body. 
\begin{equation}
    \mathcal{L}_{penetration} = \sum_k \left\Vert f(\mathbf{v}_k) \right\Vert, f(\mathbf{v}_k) < 0.
\end{equation}

Our complete objective loss function is
\begin{multline}
    \mathcal{L} = \mathcal{L}_{contact} + \lambda_1 \mathcal{L}_{normal} \\ +
    \lambda_2 \mathcal{L}_{penetration} 
    + \lambda_3 \mathcal{L}_{scale} + \lambda_4 \mathcal{L}_{reprojection}. 
    \label{eq:optmization}
\end{multline}
We experimented with various loss combinations and found $\lambda_{1,..,4}=\left[0.01, 0.01, 0.01, 0.0005\right]$ works well.

\subsection{Implementation Details}
We use GPT-3~\cite{brown2020language} as our LLM and generate semantic labels such as object sizes and human-object contact regions for reconstructing the interaction models. 
PartNet~\cite{mo2019partnet} is a large-scale dataset of 3D objects with annotated part labels. There are 24 object categories with 26.7k instances included in the dataset. We use PartNet to link the semantic labels, obtained from GPT-3, to 3D objects. A subset of shapes are selected from PartNet as representative exemplars of each category, similar to~\cite{engelmann2021points}. Each object is first centered at the origin, and we apply anisotropic scaling such that the object lies inside of a unit box. Then we compute a sign distance function for each object and discretize the unit box into $30^3$ grids, which is flattened into a vector. We use the $L_2$ distance as the similarity measure and k-Means++~\cite{arthur2006k} to cluster each category with $k=20$. We take objects that are closest to the cluster center as the representative shapes. To improve the computational speed, we downsample object meshes to 1000 vertices.

\section{Experiments}

We next provide quantitative and qualitative analysis of the proposed framework. We first evaluate the accuracy of interaction priors derived from LLMs (Sec.~\ref{sec:llmeva}) through a user study. Quantitative evaluations are carried out on the BEHAVE dataset and we compare against Phosa and ablations (Sec.~\ref{sec:behave}). We further demonstrate the generalizability of the proposed framework on new objects and interactions, and show failure cases (Sec.~\ref{sec:generalization}).   

\subsection{Evaluation of interaction priors from LLMs}
\label{sec:llmeva}
We generated semantic labels (i.e. object scales and action-conditioned human-object contact regions) for all objects from PartNet that can be detected by the detector. This results in 20 object categories.
To evaluate if the current LLMs can provide the accurate knowledge for various objects and actions, we conduct a user study to ask 10 human participants to rate if the semantic labels derived from GPT-3 are correct or not.
We consider a semantic label as \emph{correct}, \emph{uncertain}, or \emph{incorrect} based on participants' responses. 
We consider labels that have $>6$ votes as \emph{correct}, $4-6$ as \emph{uncertain}, and $<4$ as \emph{incorrect} in our analysis.
For object scales, 75\% is correct and 25\% is uncertain.
For action-conditioned contacts, 53.81\% is correct, 16.76\% is uncertain, and 29.43\% is incorrect.
Both studies have $p\ll0.001$ in binomial tests.
The semantic labels for contact regions are noisier because many of the part labels from the PartNet dataset are unfamiliar or hard to interpret for our participants.
For example, when ``holding a mug'', humans may or may not touch the ``decorations'' of the mug depending on the position of the decorations.
Overall, our user study shows that the priors derived from LLMs are mostly aligned with prior knowledge of humans about the objects and actions.

\begin{figure}
    \centering
    \includegraphics[width=0.9\linewidth]{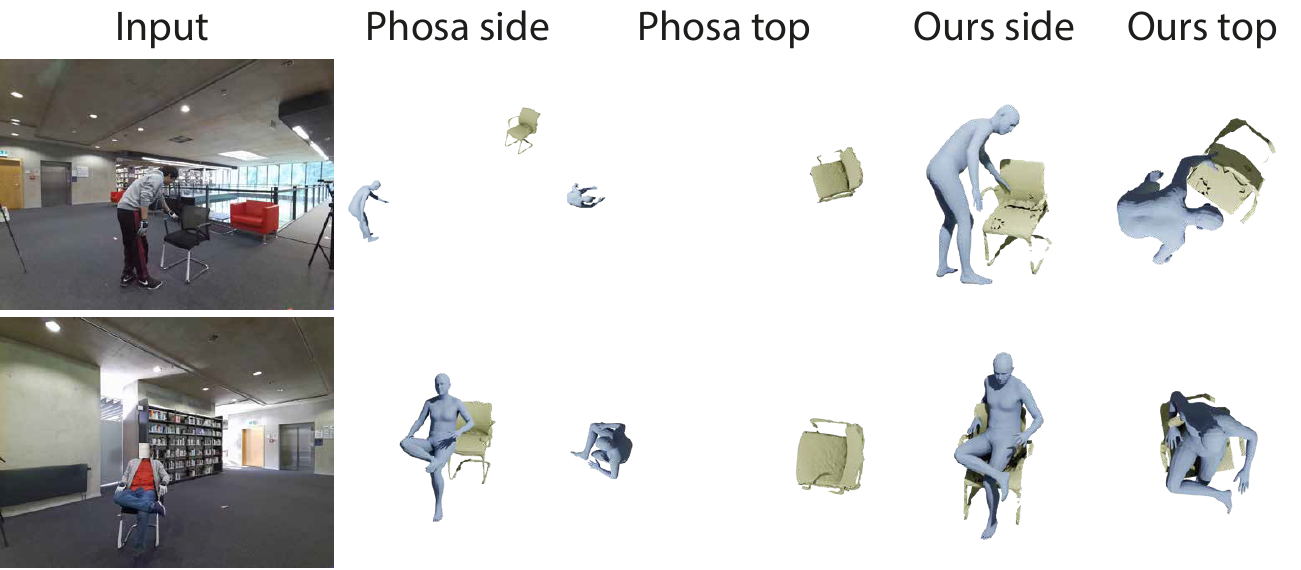}
    \caption{\textbf{Examples of the reconstruction results using BEHAVE.} Results using our method are compared against Phosa.}
    \label{fig:behave}
\end{figure}

\subsection{Quantitative evaluation of the framework}
\label{sec:behave}

\head{BEHAVE Dataset.} We evaluate our proposed method on the BEHAVE~\cite{bhatnagar2022behave} dataset, a large human-object interaction dataset containing RGB-D frames of people interacting with objects in diverse ways. BEHAVE consists of ~15k frames depicting humans interacting with 20 common objects. Accurate pseudo ground truth SMPL body models are provided for each frame, together with fitted pseudo ground truth object shapes and poses. 

\head{Experiment Setup.}
We use the test set of BEHAVE (4.5k frames) to quantitatively evaluate our approach on the following eight object categories that overlap with PartNet~\cite{mo2019partnet}: backpack, chairwood, chairblack, keyboard, stool, suitcase, tablesmall, tablesquare. Note that different interactions with the same object are considered as separate categories in BEHAVE, and we group them together. For example, backpack hold, backpack pick, and backpack lift are merged together as category backpack in our experiments. As a result, we have four main object categories, namely \textit{backpack} (merged from backpack and suitcase), \textit{chair} (merged from chairwood, chairblack and stool), \textit{keyboard} and \textit{table} (merged from tablesmall and tablesquare).

We observe that the quality of object detection from the first step largely influences the final outcome. From the four categories we test in BEHAVE, the detector fails to make a correct detection $50\%$ of the time. In all the keyboard images, only $0.5\%$ of them are correctly detected when confidence score is set to $0.9$ in PointRend~\cite{kirillov2020pointrend}. The success rate improves slightly to $2.5\%$ when we lower the confidence threshold to $0.6$. Similarly, tables are corrected detected in only $0.1\%$ (three frames) of the data and lowering the confidence score threshold does not make a difference. The detector also fails to detect the white stool in the dataset. We exclude the set of images where no objects are detected in further analysis that leaves us with the two categories \textit{chair} and \textit{backpack}. 
Given the limited interaction types of \textit{backpack} (back, hand and hug), the following analysis is focused on \textit{chair}. 

\head{Baseline.}
As an optimization-based approach, we directly reconstruct 3D human-object interaction models without additional knowledge of scene geometry or depth information. Therefore, we compare against Phosa~\cite{zhang2020perceiving}, the relevant optimization-based approach following a similar experiment setup in BEHAVE. 

\head{Evaluation Metric.}
To evaluate the reconstruction accuracy, the estimated SMPL models are first aligned to the ground truth using Procrustes analysis. We then apply the same transformation to the objects and compute the Chamfer distance for humans and objects separately.   

\begin{table}[]
    \centering
    \renewcommand{\arraystretch}{1.1}
    \setlength\tabcolsep{5.5pt}%
    \resizebox{\linewidth}{!}{%
    \small
    \begin{tabular}{lcccc}
        \hline
          \multirow{2}{3em}{Method} & \multicolumn{2}{c}{Phosa} & \multicolumn{2}{c}{Ours} \\
          & $\mathcal{H}$ $\downarrow$ & $\mathcal{O}$ $\downarrow$  & $\mathcal{H}$ $\downarrow$ & $\mathcal{O}$$\downarrow$  \\
          \hline
          
          Hand & 7.4 (2.5) & 83.3 (167.7) & 8.1 (2.6)	& 34.3 (17.6) \\
          Lift & 8.0 (2.9) & 88.0 (189.0) & 8.3	(3.1) & 29.1 (16.7) \\
          Liftreal & 7.7 (2.2) & 80.1 (172.8) & 7.8 (2.4)	& 29.1 (14.3) \\
          Sit & 7.1 (2.5) & 32.2 (57.6) & 7.9 (3.1) & 23.4 (12.5) \\
          Sitstand & 6.3 (1.5) & 26.9 (8.2) &8.1 (2.1) &	25.1 (12.7) \\
          Mix & 6.5 (2.3) & 88.1 (187.7) & 7.4 (3.0) & 27.5 (16.7) \\
          \hdashline
          Avg. & 7.2 (2.3) & 66.4 (130.5) & 7.9 (2.7) & 28.1 (15.1)\\
         \hline
    \end{tabular}
    }
    \vspace{0.1em}
    \caption{\textbf{BEHAVE Chair.} We compare our methods for inferring 3D humans and objects from images to Phosa~\cite{speer2017conceptnet} on the chair category from the BEHAVE dataset. We distinguish between different actions. Chamfer distance is used to evaluate the reconstructed SMPL models $\mathcal{H}$ and 3D objects $\mathcal{O}$ independently. Mean and standard deviation are reported in cm.}
    \label{tab:reconstruction}
\end{table}

\head{Reconstruction accuracy.} 
We show two examples of our method and Phosa for BEHAVE images in Figure~\ref{fig:behave}. Without the ability to adjust the way of contact between humans and objects, Phosa struggles with obtaining meaningful 3D models and very often results in situations where humans and objects do not contact.
We notice that small objects like backpacks are mostly limited to hand-object interactions.  
Therefore, we only quantify the results on chairs in Table~\ref{tab:reconstruction} and compare to the pseudo ground truth provided in BEHAVE. We report reconstruction accuracy for estimated humans $\mathcal{H}$ and objects $\mathcal{O}$ separately and distinguish between different actions. Note that action types are defined by BEHAVE. 

Our estimated humans are on par with the estimated humans from Phosa, even though we have slightly larger average errors. However, Phosa does not consider different action types and performs badly when it comes to different interactions, resulting in large errors in reconstructed objects, see Table~\ref{tab:reconstruction}. Additionally, we use ground truth object masks provided by BEHAVE and conduct more quantitative evaluations on the remaining three object object categories in the Supplementary Material. 

\begin{table}[t]
    \centering
    \centering
    \renewcommand{\arraystretch}{1.1}  \setlength\tabcolsep{5.5pt}%
    \small
    \begin{tabular}{lcc}
        \hline
         \textbf{Method} & $\mathcal{H}$ & $\mathcal{O}$ \\
         \hline
         No action & 7.9 (2.9) & 58.4 (40.3)\\
         No $\mathcal{L}_{contact}$ & 8.0 (2.7)	& 103.2	(86.3)\\
         No $\mathcal{L}_{normal}$ & 7.8 (2.8) & 30.7 (14.3) \\
         Ours (full) & 7.9 (2.7) & 28.1 (15.1)\\
         \hline
    \end{tabular} 
    \vspace{0.2em}
    \caption{\textbf{Ablation.} We investigate how different steps impact the final reconstruction accuracy and ablate the core components.}
    \label{tab:ablation}
\end{table}

\begin{figure}
    \centering
    \includegraphics[width=0.9\linewidth]{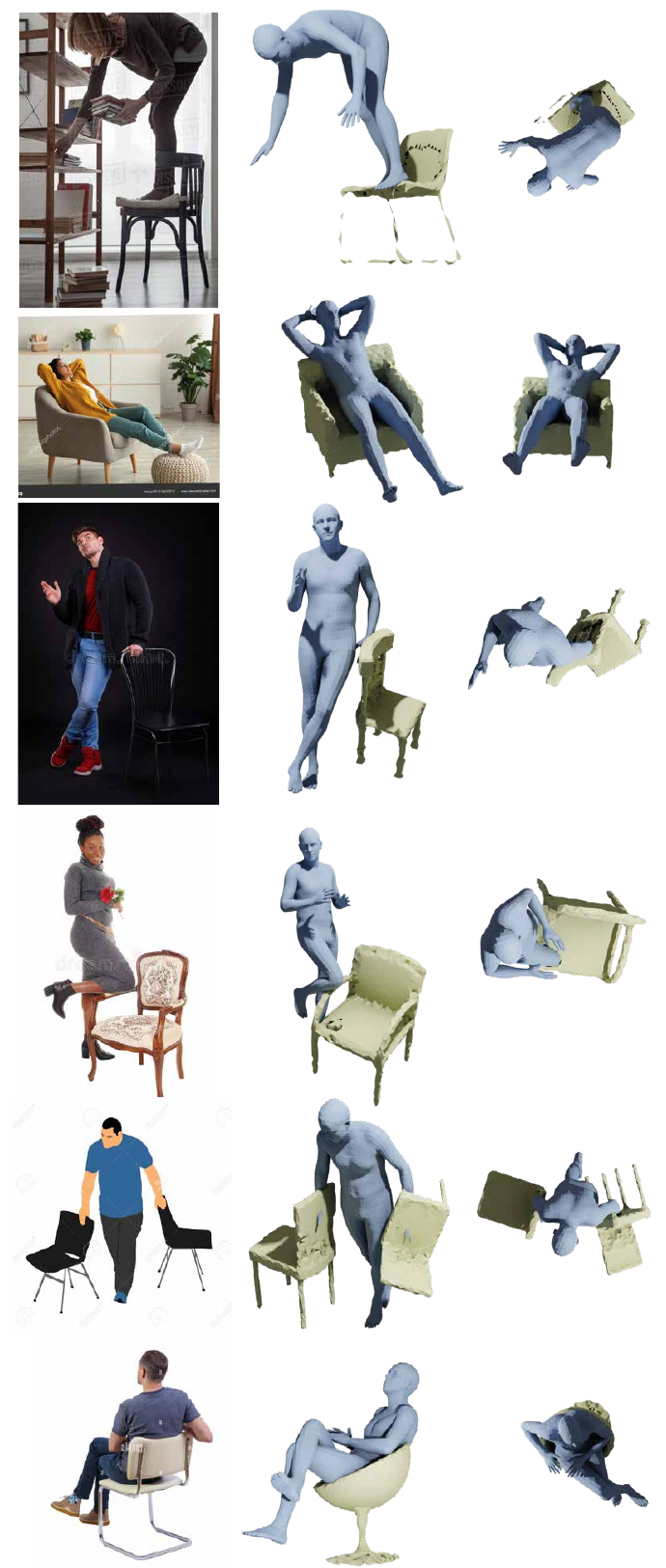}
    \caption{\textbf{Examples of inferred 3D human-chair models on various interactions.} Our method can reconstruct 3D models of different interactions that require various types of contacts.}
    \label{fig:chair}
\end{figure}

\begin{figure*}
    \centering
    \includegraphics[width=0.9\linewidth]{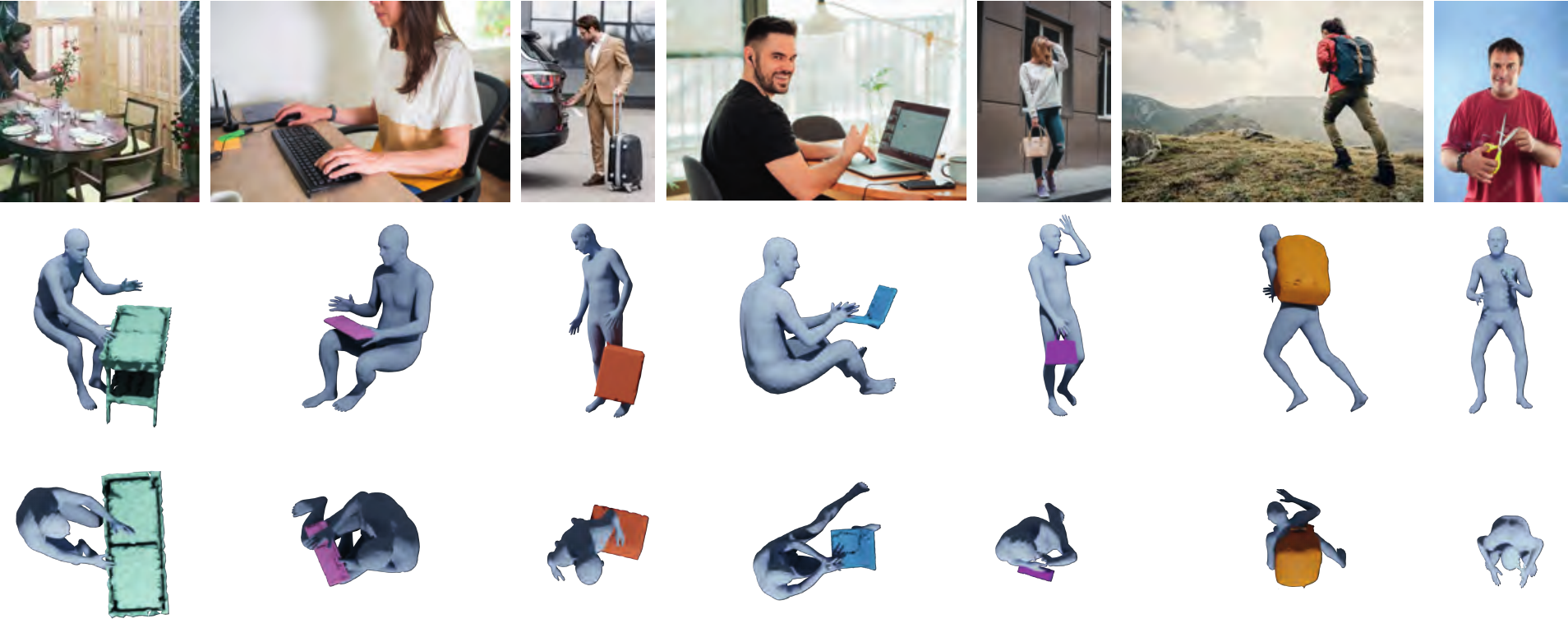}
    \caption{\textbf{Examples of inferred 3D interaction models on various object categories.} Using the proposed optimization approach, we can reconstruct diverse interaction models for PartNet objects.}
    \label{fig:generalize}
\end{figure*}

\head{Ablation.}
In Table~\ref{tab:ablation}, we ablate the proposed method to analyze its core components: \textit{No action} ignores the underlying action type and directly infers the 3D models using the default contact configuration, \textit{No $\mathcal{L}_{contact}$} does not use the contact loss in joint optimization and \textit{No $\mathcal{L}_{normal}$} does not consider normal directions when optimizing contacts between humans and objects. 
Interaction with the largest weight in ConceptNet is used as the default contact. 
The performance on estimated humans does not change significantly, 
however, without $\mathcal{L}_{contact}$, the accuracy of obtained objects drops significantly. And $\mathcal{L}_{normal}$ has rather limited contribution on average.

\begin{figure*}[b]
    \centering
    \includegraphics[width=0.9\textwidth]{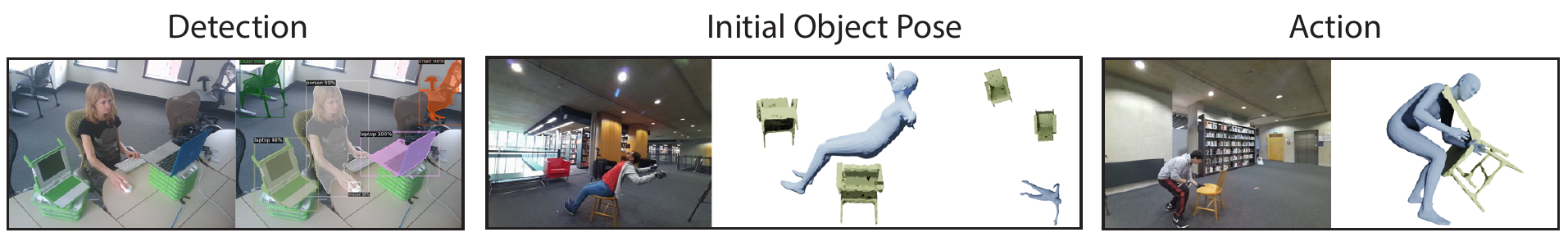}
    \caption{\textbf{Failure cases.} Our method fails to infer valid interaction models when objects are not detected (Detection), their poses are wrongly initialized (Initial Object Pose), or the recognised action type is incorrect (Action).}
    \label{fig:failure}
\end{figure*}

\subsection{Generalization to new objects and interactions}
\label{sec:generalization}
Using knowledge priors extracted from LLMs, the proposed method can be applied to new objects and interactions. To show our method can generalize, we reconstruct various human-object interaction models for PartNet objects. Figure~\ref{fig:chair} shows different examples of human interacting with a chair.
We observe that our method can successfully model interactions that require different types of contacts. 
A qualitative analysis of the reconstruction results of different objects (examples shown in Figure~\ref{fig:generalize}) shows that the proposed method can generalize towards a wide range of object categories (more examples in the Supp. Mat.).  
We would like to highlight that existing approaches require expensive, method-specific human annotations, which poses a severe limitation on the diversity of object categories. We provide a method to bring in existing annotations and data. 

\vspace{1mm}
\head{Failure cases.} Figure~\ref{fig:failure} shows some typical failure cases. 
Our method fails to infer meaningful 3D models when objects are not well detected (Detection in Figure~\ref{fig:failure}). We fail to obtain valid interaction models when object poses are poorly  initialized (Initial Object Pose in Figure~\ref{fig:failure}). 
In addition, we also observe that retrieval-based action recognition may fail to identify the true action type. In the Action example in Figure~\ref{fig:failure}, a person is touching a chair while forming a pose that is similar to sitting poses. 
More sophisticated approaches that take image content of both humans and objects into account could address such failure cases and may provide more accurate contact information.

\section{Discussion}

Reconstructing 3D human and object models from images is a crucial step for interaction modeling. We propose a pipeline that leverages commonsense knowledge extracted from LLMs to solve low-level computer vision tasks. It allows us to model different interactions without the constraint post by human annotations. 
Off-the-shelf computer vision algorithms are used to detect object and estimate the initial object shape and pose. We expect improvement with better methods available for these subtasks. While we rely on the PartNet dataset, we also expect progress on object part segmentation tasks. We believe our work provides a useful new way to model human-object interactions.  

\vspace{1mm}
\head{Limitations.}
The proposed approach leaves ample room for future work. Our current prompt design needs to be upgraded to capture diverse interactions under the same action condition. For example, while carrying a chair, one can hold on the seat or the back. The pose-based action recognition is reasonable for full-body interactions with large objects, but handhold object interactions demand fine-grained contact information of hands. The reconstruction pipeline can largely benefit from better initialization of object poses and using richer image evidence can certainly improve the quality of reconstructed interaction models. 

\section*{Acknowledgement}    
This work was supported by an ETH Zurich Postdoctoral Fellowship.
We thank the anonymous reviewers and all participants of our user study. 
We are grateful to Xu Chen and Ronny H\"ansch for helpful discussions and comments. 

{\small
\bibliographystyle{ieee_fullname}
\bibliography{egbib}

\begin{thebibliography}{10}\itemsep=-1pt

\bibitem{akula2021robust}
Arjun Akula, Varun Jampani, Soravit Changpinyo, and Song-Chun Zhu.
\newblock Robust visual reasoning via language guided neural module networks.
\newblock {\em Advances in Neural Information Processing Systems}, 34, 2021.

\bibitem{arthur2006k}
David Arthur and Sergei Vassilvitskii.
\newblock k-means++: The advantages of careful seeding.
\newblock Technical report, Stanford, 2006.

\bibitem{bhatnagar2022behave}
Bharat~Lal Bhatnagar, Xianghui Xie, Ilya~A Petrov, Cristian Sminchisescu,
  Christian Theobalt, and Gerard Pons-Moll.
\newblock Behave: Dataset and method for tracking human object interactions.
\newblock {\em arXiv preprint arXiv:2204.06950}, 2022.

\bibitem{brown2020language}
Tom Brown, Benjamin Mann, Nick Ryder, Melanie Subbiah, Jared~D Kaplan, Prafulla
  Dhariwal, Arvind Neelakantan, Pranav Shyam, Girish Sastry, Amanda Askell,
  et~al.
\newblock Language models are few-shot learners.
\newblock {\em Advances in neural information processing systems},
  33:1877--1901, 2020.

\bibitem{bucker2022reshaping}
Arthur Bucker, Luis Figueredo, Sami Haddadin, Ashish Kapoor, Shuang Ma, and
  Rogerio Bonatti.
\newblock Reshaping robot trajectories using natural language commands: A study
  of multi-modal data alignment using transformers.
\newblock {\em arXiv preprint arXiv:2203.13411}, 2022.

\bibitem{mmpose2020}
MMPose Contributors.
\newblock Openmmlab pose estimation toolbox and benchmark.
\newblock \url{https://github.com/open-mmlab/mmpose}, 2020.

\bibitem{engelmann2021points}
Francis Engelmann, Konstantinos Rematas, Bastian Leibe, and Vittorio Ferrari.
\newblock From points to multi-object 3d reconstruction.
\newblock In {\em Proceedings of the IEEE/CVF Conference on Computer Vision and
  Pattern Recognition}, pages 4588--4597, 2021.

\bibitem{gavrila2000pedestrian}
Dariu~M Gavrila.
\newblock Pedestrian detection from a moving vehicle.
\newblock In {\em European conference on computer vision}, pages 37--49.
  Springer, 2000.

\bibitem{hassan2019resolving}
Mohamed Hassan, Vasileios Choutas, Dimitrios Tzionas, and Michael~J Black.
\newblock Resolving 3d human pose ambiguities with 3d scene constraints.
\newblock In {\em Proceedings of the IEEE/CVF international conference on
  computer vision}, pages 2282--2292, 2019.

\bibitem{henzler2019escaping}
Philipp Henzler, Niloy~J Mitra, and Tobias Ritschel.
\newblock Escaping plato's cave: 3d shape from adversarial rendering.
\newblock In {\em Proceedings of the IEEE/CVF International Conference on
  Computer Vision}, pages 9984--9993, 2019.

\bibitem{huang2022language}
Wenlong Huang, Pieter Abbeel, Deepak Pathak, and Igor Mordatch.
\newblock Language models as zero-shot planners: Extracting actionable
  knowledge for embodied agents.
\newblock {\em arXiv preprint arXiv:2201.07207}, 2022.

\bibitem{jones1991object}
Susan~S Jones, Linda~B Smith, and Barbara Landau.
\newblock Object properties and knowledge in early lexical learning.
\newblock {\em Child development}, 62(3):499--516, 1991.

\bibitem{kato2018neural}
Hiroharu Kato, Yoshitaka Ushiku, and Tatsuya Harada.
\newblock Neural 3d mesh renderer.
\newblock In {\em Proceedings of the IEEE conference on computer vision and
  pattern recognition}, pages 3907--3916, 2018.

\bibitem{kirillov2020pointrend}
Alexander Kirillov, Yuxin Wu, Kaiming He, and Ross Girshick.
\newblock Pointrend: Image segmentation as rendering.
\newblock In {\em Proceedings of the IEEE/CVF conference on computer vision and
  pattern recognition}, pages 9799--9808, 2020.

\bibitem{Kocabas_PARE_2021}
Muhammed Kocabas, Chun-Hao~P. Huang, Otmar Hilliges, and Michael~J. Black.
\newblock {PARE}: Part attention regressor for {3D} human body estimation.
\newblock In {\em Proceedings International Conference on Computer Vision
  (ICCV)}, pages 11127--11137. IEEE, Oct. 2021.

\bibitem{kocabas2021spec}
Muhammed Kocabas, Chun-Hao~P Huang, Joachim Tesch, Lea M{\"u}ller, Otmar
  Hilliges, and Michael~J Black.
\newblock Spec: Seeing people in the wild with an estimated camera.
\newblock In {\em Proceedings of the IEEE/CVF International Conference on
  Computer Vision}, pages 11035--11045, 2021.

\bibitem{kuo:compositional-gscan2021}
Yen-Ling Kuo, Boris Katz, and Andrei Barbu.
\newblock Compositional networks enable systematic generalization for grounded
  language understanding.
\newblock In {\em Findings of the Association for Computational Linguistics:
  EMNLP 2021}, 2021.

\bibitem{li2002turning}
Peggy Li and Lila Gleitman.
\newblock Turning the tables: Language and spatial reasoning.
\newblock {\em Cognition}, 83(3):265--294, 2002.

\bibitem{liao2019natural}
Wentong Liao, Bodo Rosenhahn, Ling Shuai, and Michael Ying~Yang.
\newblock Natural language guided visual relationship detection.
\newblock In {\em Proceedings of the IEEE/CVF Conference on Computer Vision and
  Pattern Recognition Workshops}, 2019.

\bibitem{SMPL:2015}
Matthew Loper, Naureen Mahmood, Javier Romero, Gerard Pons-Moll, and Michael~J.
  Black.
\newblock {SMPL}: A skinned multi-person linear model.
\newblock {\em ACM Trans. Graphics (Proc. SIGGRAPH Asia)}, 34(6):248:1--248:16,
  Oct. 2015.

\bibitem{mo2019partnet}
Kaichun Mo, Shilin Zhu, Angel~X Chang, Li Yi, Subarna Tripathi, Leonidas~J
  Guibas, and Hao Su.
\newblock Partnet: A large-scale benchmark for fine-grained and hierarchical
  part-level 3d object understanding.
\newblock In {\em Proceedings of the IEEE/CVF conference on computer vision and
  pattern recognition}, pages 909--918, 2019.

\bibitem{narasimhan2021clip}
Medhini Narasimhan, Anna Rohrbach, and Trevor Darrell.
\newblock Clip-it! language-guided video summarization.
\newblock {\em Advances in Neural Information Processing Systems}, 34, 2021.

\bibitem{niemeyer2020differentiable}
Michael Niemeyer, Lars Mescheder, Michael Oechsle, and Andreas Geiger.
\newblock Differentiable volumetric rendering: Learning implicit 3d
  representations without 3d supervision.
\newblock In {\em Proceedings of the IEEE/CVF Conference on Computer Vision and
  Pattern Recognition}, pages 3504--3515, 2020.

\bibitem{patki2020language}
Siddharth Patki, Ethan Fahnestock, Thomas~M Howard, and Matthew~R Walter.
\newblock Language-guided semantic mapping and mobile manipulation in partially
  observable environments.
\newblock In {\em Conference on Robot Learning}, pages 1201--1210. PMLR, 2020.

\bibitem{SMPL-X:2019}
Georgios Pavlakos, Vasileios Choutas, Nima Ghorbani, Timo Bolkart, Ahmed A.~A.
  Osman, Dimitrios Tzionas, and Michael~J. Black.
\newblock Expressive body capture: {3D} hands, face, and body from a single
  image.
\newblock In {\em Proceedings IEEE Conf. on Computer Vision and Pattern
  Recognition (CVPR)}, pages 10975--10985, 2019.

\bibitem{petroni-etal-2019-language}
Fabio Petroni, Tim Rockt{\"a}schel, Sebastian Riedel, Patrick Lewis, Anton
  Bakhtin, Yuxiang Wu, and Alexander Miller.
\newblock Language models as knowledge bases?
\newblock In {\em Proceedings of the 2019 Conference on Empirical Methods in
  Natural Language Processing and the 9th International Joint Conference on
  Natural Language Processing (EMNLP-IJCNLP)}, pages 2463--2473, Hong Kong,
  China, Nov. 2019. Association for Computational Linguistics.

\bibitem{radford2021learning}
Alec Radford, Jong~Wook Kim, Chris Hallacy, Aditya Ramesh, Gabriel Goh,
  Sandhini Agarwal, Girish Sastry, Amanda Askell, Pamela Mishkin, Jack Clark,
  et~al.
\newblock Learning transferable visual models from natural language
  supervision.
\newblock In {\em International Conference on Machine Learning}, pages
  8748--8763. PMLR, 2021.

\bibitem{rao2021denseclip}
Yongming Rao, Wenliang Zhao, Guangyi Chen, Yansong Tang, Zheng Zhu, Guan Huang,
  Jie Zhou, and Jiwen Lu.
\newblock Denseclip: Language-guided dense prediction with context-aware
  prompting.
\newblock In {\em Proceedings of the IEEE Conference on Computer Vision and
  Pattern Recognition (CVPR)}, 2022.

\bibitem{richter2018matryoshka}
Stephan~R Richter and Stefan Roth.
\newblock Matryoshka networks: Predicting 3d geometry via nested shape layers.
\newblock In {\em Proceedings of the IEEE conference on computer vision and
  pattern recognition}, pages 1936--1944, 2018.

\bibitem{MANO:SIGGRAPHASIA:2017}
Javier Romero, Dimitrios Tzionas, and Michael~J. Black.
\newblock Embodied hands: Modeling and capturing hands and bodies together.
\newblock {\em ACM Transactions on Graphics, (Proc. SIGGRAPH Asia)}, 36(6),
  2017.

\bibitem{savva2016pigraphs}
Manolis Savva, Angel~X Chang, Pat Hanrahan, Matthew Fisher, and Matthias
  Nie{\ss}ner.
\newblock Pigraphs: learning interaction snapshots from observations.
\newblock {\em ACM Transactions on Graphics (TOG)}, 35(4):1--12, 2016.

\bibitem{sharma2022skill}
Pratyusha Sharma, Antonio Torralba, and Jacob Andreas.
\newblock Skill induction and planning with latent language.
\newblock In {\em ACL 2022}, May 2022.

\bibitem{speer2017conceptnet}
Robyn Speer, Joshua Chin, and Catherine Havasi.
\newblock Conceptnet 5.5: An open multilingual graph of general knowledge.
\newblock In {\em Thirty-first AAAI conference on artificial intelligence},
  2017.

\bibitem{starke2019neural}
Sebastian Starke, He Zhang, Taku Komura, and Jun Saito.
\newblock Neural state machine for character-scene interactions.
\newblock {\em ACM Trans. Graph.}, 38(6):209--1, 2019.

\bibitem{taylor1953cloze}
Wilson~L Taylor.
\newblock “cloze procedure”: A new tool for measuring readability.
\newblock {\em Journalism quarterly}, 30(4):415--433, 1953.

\bibitem{weng2021holistic}
Zhenzhen Weng and Serena Yeung.
\newblock Holistic 3d human and scene mesh estimation from single view images.
\newblock In {\em Proceedings of the IEEE/CVF Conference on Computer Vision and
  Pattern Recognition}, pages 334--343, 2021.

\bibitem{xie2022chore}
Xianghui Xie, Bharat~Lal Bhatnagar, and Gerard Pons-Moll.
\newblock Chore: Contact, human and object reconstruction from a single rgb
  image.
\newblock {\em arXiv preprint arXiv:2204.02445}, 2022.

\bibitem{yi2022human}
Hongwei Yi, Chun-Hao~P Huang, Dimitrios Tzionas, Muhammed Kocabas, Mohamed
  Hassan, Siyu Tang, Justus Thies, and Michael~J Black.
\newblock Human-aware object placement for visual environment reconstruction.
\newblock {\em arXiv preprint arXiv:2203.03609}, 2022.

\bibitem{zeng2022socratic}
Andy Zeng, Adrian Wong, Stefan Welker, Krzysztof Choromanski, Federico Tombari,
  Aveek Purohit, Michael Ryoo, Vikas Sindhwani, Johnny Lee, Vincent Vanhoucke,
  et~al.
\newblock Socratic models: Composing zero-shot multimodal reasoning with
  language.
\newblock {\em arXiv preprint arXiv:2204.00598}, 2022.

\bibitem{zhang2020perceiving}
Jason~Y Zhang, Sam Pepose, Hanbyul Joo, Deva Ramanan, Jitendra Malik, and
  Angjoo Kanazawa.
\newblock Perceiving 3d human-object spatial arrangements from a single image
  in the wild.
\newblock In {\em European conference on computer vision}, pages 34--51.
  Springer, 2020.

\bibitem{zhang2022couch}
Xiaohan Zhang, Bharat~Lal Bhatnagar, Vladimir Guzov, Sebastian Starke, and
  Gerard Pons-Moll.
\newblock Couch: Towards controllable human-chair interactions.
\newblock {\em arXiv preprint arXiv:2205.00541}, 2022.

\end{thebibliography}
}

\clearpage
\appendix

\section{ Supplemental Material}
\subsection{Prompt templates}
\label{app:prompt}
We use fill-in-the-blank prompts to query LLMs and below are the prompt templates we used to query GPT-3.

\paragraph{Object size.} We use the following prompt template to obtain object size by replacing OBJECT with a specific object category. 
\begin{verbatim}
This is an object length estimator.
Length of a bike: 1.75m
Height of a woman: 1.63m
Length of a OBJECT:
\end{verbatim}

\paragraph{Contact regions of body and object parts.} We define interactions as pairs of ACTION and OBJECT and use the following template to obtain part-level body-object contacts. While designing the prompt, we noticed that it is important to include different interactions for one object category (e.g. ride and walk bike) and have examples of different body parts. We also noticed that it is better to use simple words for body parts (e.g. using back instead of spine). 

\begin{verbatim}
This is a body-object contact generator.

Action: ride
Object: bike
Contacts: handlebar/hands, seat/butt, 
          paddle/foot

Action: walk 
Object: bike
Contacts: handlebar/hands

Action: sit
Object: sofa
Contacts: seat/butt, back/back

Action: stand
Object: sofa
Contacts: seat/foot

Action: kick
Object: soccer
Contacts: soccer/foot

Action: carry
Object: soccer
Contacts: soccer/hands

Action: ACTION
Object: OBJECT
Contacts:
\end{verbatim}

\begin{table}[]
    \centering
    \begin{tabular}{c|c|c|c}
        \hline
        Category & Size & Category & Size  \\
        \hline
        Backpack & 0.5 & Bag & 0.5 \\ 
        Bed & 2.0 & Bottle & 0.3 \\
        Bowl	& 0.15 & Chair	& 0.85 \\
        Clock & 0.3 & Couch & 0.91 \\ 
        Cup & 0.1 & Desk & 0.75 \\
        Door & 2.1 & Handbag & 0.3 \\
        Hat & 	0.3 & Keyboard	& 0.61 \\
        Knife &	0.22 & Microwave &	0.5 \\ 
        Mug	& 0.12 & Scissors	& 0.2 \\
        Suitcase & 0.81 & Table & 	0.75 \\
        \hline
    \end{tabular}
    \vspace{0.3em}
    \caption{\textbf{Size of objects obtained from LLMs.} The obtained object size is reported in meters. We use them as the initialization for object scale in the joint optimization.}
    \label{tab:objsize}
\end{table}

\begin{table}
    \centering
    \begin{tabular}{c|c|c}
        \hline
        Category & Action & Contact (Object Part, Body Part)\\
        \hline
        \multirow{11}{3em}{Chair} & sit & (chair seat, butt),\\
        && (chair back, back)\\\cline{2-3}
        & carry & (chair arms, hands), \\
        && (chair back, hands), \\ 
        && (chair seat, hands) \\\cline{2-3}
        & rest  & (chair seat, butt),\\
        && (chair back, back)\\\cline{2-3}
        & stand on & (chair seat, feet)\\\cline{2-3}
        & stand next to & (chair back, hands)\\\cline{2-3}
        & sleep  & (chair seat, butt),\\
        && (chair back, back)\\
        \hline
        \multirow{7}{3em}{Table} & sit & (tabletop, butt) \\
        && (tabletop, left leg) \\
        && (tabletop, right leg) \\\cline{2-3}
        & work & (tabletop, hands) \\\cline{2-3}
        & arrange & (tabletop, hands) \\\cline{2-3}
        & lay & (tabletop, body) \\\cline{2-3}
        & place & (tabletop, hands) \\
        \hline 
        \multirow{9}{4em}{Backpack} & carry & (shoulder strap, hands) \\
        && (support, hands) \\\cline{2-3}
        & backpack & (shoulder strap, shoulders) \\
        && (support, shoulders) \\
        && (bag body, back) \\\cline{2-3}
        & mount & (shoulder strap, hands)\\ 
        && (shoulder strap, waist)\\ 
        && (support, hands)\\
        && (support, waist)\\
        \hline
        \multirow{3}{4em}{Suitcase} & carry & (handle, hands) \\\cline{2-3}
        & pack & (zipper, hands) \\\cline{2-3}
        & lug & (handle, hands) \\ 
        & throw & (handle, hands) \\
        \hline
        \multirow{6}{4em}{Scissors} & cut & (blade handle, hands) \\
        && (handle, hands) \\\cline{2-3}
        & pass & (blade, hands) \\
        && (blade handle, hands) \\
        && (handle, hands) \\
        && (securing clip, hands) \\
        \hline         
        \multirow{4}{4em}{Keyboard} & type & (key, hands) \\\cline{2-3}
        & play & (key, hands) \\\cline{2-3}
        & control & (key, hands) \\\cline{2-3}
        & enter & (key, hands) \\
        \hline
        \multirow{4}{3em}{Bowl} & hold & (bowl, hands) \\\cline{2-3}
        & serve & (bowl, hands) \\\cline{2-3}
        & eat & (bowl, mouth) \\\cline{2-3}
        & wash & (bowl, hands) \\
        \hline
    \end{tabular}
    \vspace{0.3em}
    \caption{\textbf{Example of contact parts obtained through LLM query.} Given an object category and an action, we query LLMs for part-level contacts. Each parts pair consists of an object part and a body part. We use PartNet labels as object part labels, and SMPL labels for body part labels. These parts pairs are used in $\mathcal{L}_{contact}$ and $\mathcal{L}_{normal}$ to bring parts pairs closer to each other.}
    \label{tab:contactexample}
\end{table}

\subsection{Example semantic labels from LLMs}
\label{app:llm_examples}

Table~\ref{tab:objsize} shows a list of obtained object size in meters and examples of human-object contact regions can be found in Table~\ref{tab:contactexample}. For small objects, their interactions are mostly limited to hand-object interactions instead of body-object interactions.   

\subsection{Preprocessing PartNet objects}
PartNet provides object instances, which has 10,000 vertices each stored in point clouds, and semantic labels of parts. To have the optimization run more efficiently, we first preprocess PartNet point clouds such that each model has less vertices and faces. More specifically, we first downsample the point clouds to have 1000 vertices, and then use the Ball-Pivoting Algorithm to reconstruct surfaces. Finally, we fill the holes and remove non manifold vertices and faces using MeshLab.
Fig.~\ref{fig:add2} shows the selected 20 representative objects for the chair category.

\begin{figure}
    \centering
    \includegraphics[width=0.95\linewidth]{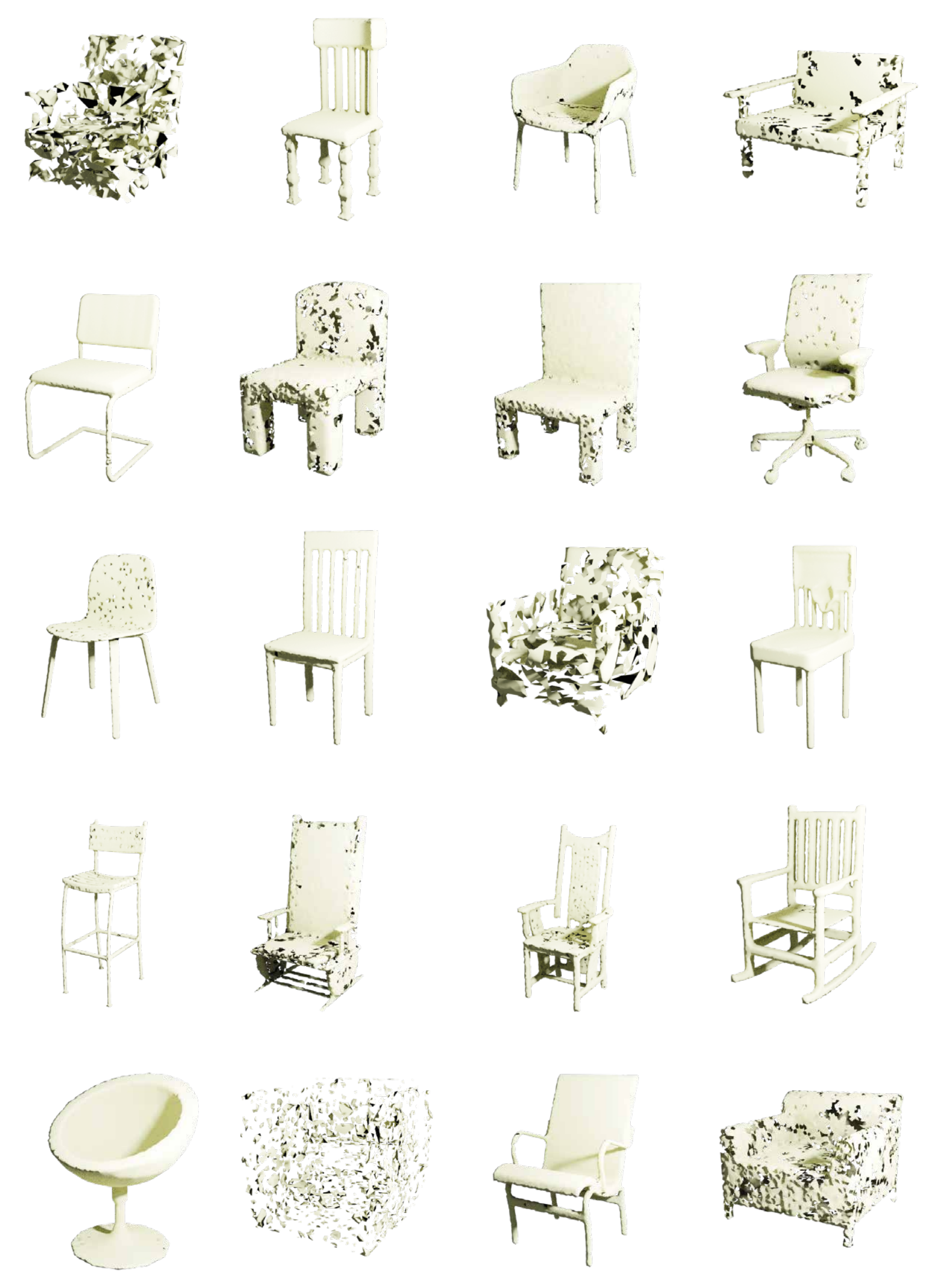}
    \caption{\textbf{20 representative objects for the chair category.}}
    \label{fig:add2}
\end{figure}

\subsection{Optimization details}
We use a differentiable renderer to estimate 6-DoF object poses using the ADAM optimizer with learning rate 1e-3 for 50 iterations. The overall objective function in Eq.~\ref{eq:optmization} is optimized with 500 ADAM updates and learning rate of 2e-3. The whole optimization takes roughly half an hour for an image.   

\subsection{Additional quantitative results}

While object detector often fails to detect other objects (\textit{table, keyboard, backpack}) in BEHAVE, their object masks are provided by the dataset. We conduct additional quantitative results using the provided ground truth object masks and evaluate on the remaining three object categories. 
Similar performance is observed in Tab.~\ref{tab:eva} where our method significantly outperforms SotA on the BEHAVE \textit{table}.
Average improvement from 298cm to 80cm for BEHAVE \textit{keyboard} (see Tab.~\ref{tab:keyboard}) and 80cm to 67cm for BEHAVE \textit{backpack} (see Tab.~\ref{tab:backpack}). 
This provides further evidence of our novel integration of knowledge priors

\begin{table}[]
    \centering
    \renewcommand{\arraystretch}{1.1}
    \setlength\tabcolsep{5.5pt}%
    \resizebox{\linewidth}{!}{%
    \small
    \begin{tabular}{lcccc}
        \hline
          \multirow{2}{3em}{Method} & \multicolumn{2}{c}{Phosa} & \multicolumn{2}{c}{Ours} \\
          & $\mathcal{H}$ $\downarrow$ & $\mathcal{O}$ $\downarrow$  & $\mathcal{H}$ $\downarrow$ & $\mathcal{O}$$\downarrow$  \\
          \hline
          Hand & 6.3 (2.0) & 621.6 (423.2) & 8.1 (2.6)	& 34.3 (17.6) \\
          Lift & 9.1 (3.8) & 528.6 (447.3) & 11.0	(2.8) & 342.1 (354.9) \\
          Move & 8.9 (3.3) & 528.8 (420.0) & 10.0 (4.6)	& 276.7 (271.0) \\
          Sit & 9.3 (4.6) & 740.4 (318.6) & 8.9 (4.7) & 161.0 (246.3) \\
          Mix & 9.2 (3.8) & 796.6 (328.7) & 9.1 (3.6) & 424.0 (3.36) \\
          \hdashline
          Avg. & 8.6 (3.5) & 643.2 (387.6) & 9.2 (3.4) & 307.9 (295.4)\\
         \hline
    \end{tabular}
    }
    \vspace{0.1em}
    \caption{\textbf{BEHAVE \textit{table}.} We compare our methods for inferring 3D humans and objects from images to Phosa~\cite{speer2017conceptnet} on the \textit{table} category from the BEHAVE dataset. We distinguish between different actions. Chamfer distance is used to evaluate the reconstructed SMPL models $\mathcal{H}$ and 3D objects $\mathcal{O}$ independently. Mean and standard deviation are reported in cm.}
    \label{tab:eva}
    \vspace{-1em}
\end{table}

\begin{table}[]
    \centering
    \renewcommand{\arraystretch}{1.1}
    \setlength\tabcolsep{5.5pt}%
    \resizebox{\linewidth}{!}{%
    \small
    \begin{tabular}{lcccc}
        \hline
          \multirow{2}{3em}{Method} & \multicolumn{2}{c}{Phosa} & \multicolumn{2}{c}{Ours} \\
          & $\mathcal{H}$ $\downarrow$ & $\mathcal{O}$ $\downarrow$  & $\mathcal{H}$ $\downarrow$ & $\mathcal{O}$$\downarrow$  \\
          \hline
          Move & 8.3 (1.9) & 168.9 (325.2) & 7.8 (1.9)	& 68.6 (162.7) \\
          Type & 6.2 (2.0) & 426.6 (440.1) & 5.1 (1.4) & 91.3 (257.2) \\
          \hdashline
          Avg. & 7.2 (1.9) & 297.7 (382.7) & 6.5 (1.6) & 80.0 (209.9)\\
         \hline
    \end{tabular}
    }
    \vspace{0.1em}
    \caption{\textbf{BEHAVE \textit{keyboard}.} We compare our methods for inferring 3D humans and objects from images to Phosa~\cite{speer2017conceptnet} on the \textit{keyboard} category from the BEHAVE dataset. We distinguish between different actions. Chamfer distance is used to evaluate the reconstructed SMPL models $\mathcal{H}$ and 3D objects $\mathcal{O}$ independently. Mean and standard deviation are reported in cm.}
    \label{tab:keyboard}
    \vspace{-1em}
\end{table}

\begin{table}[]
    \centering
    \renewcommand{\arraystretch}{1.1}
    \setlength\tabcolsep{5.5pt}%
    \resizebox{\linewidth}{!}{%
    \small
    \begin{tabular}{lcccc}
        \hline
          \multirow{2}{3em}{Method} & \multicolumn{2}{c}{Phosa} & \multicolumn{2}{c}{Ours} \\
          & $\mathcal{H}$ $\downarrow$ & $\mathcal{O}$ $\downarrow$  & $\mathcal{H}$ $\downarrow$ & $\mathcal{O}$$\downarrow$  \\
          \hline
          Hand & 7.2 (2.5) & 132.0 (234.2) & 8.5 (2.6)	& 84.9 (161.1) \\
          Hug & 7.7 (3.0) & 166.5 (221.4) & 7.8	(2.4) & 19.5 (127.3) \\
          Back & 6.0 (2.0) & 92.6 (218.6) & 6.7 (1.5)	& 97.2 (228.1) \\
          \hdashline
          Avg. & 7.0 (2.5) & 80.4 (224.7) & 7.7 (2.2) & 67.2 (172.2)\\
         \hline
    \end{tabular}
    }
    \vspace{0.1em}
    \caption{\textbf{BEHAVE \textit{backpack}.} We compare our methods for inferring 3D humans and objects from images to Phosa~\cite{speer2017conceptnet} on the \textit{backpack} category from the BEHAVE dataset. We distinguish between different actions. Chamfer distance is used to evaluate the reconstructed SMPL models $\mathcal{H}$ and 3D objects $\mathcal{O}$ independently. Mean and standard deviation are reported in cm.}
    \label{tab:backpack}
    \vspace{-1em}
\end{table}

\subsection{Additional qualitative results}

\begin{figure*}
    \centering
    \includegraphics[width=0.95\linewidth]{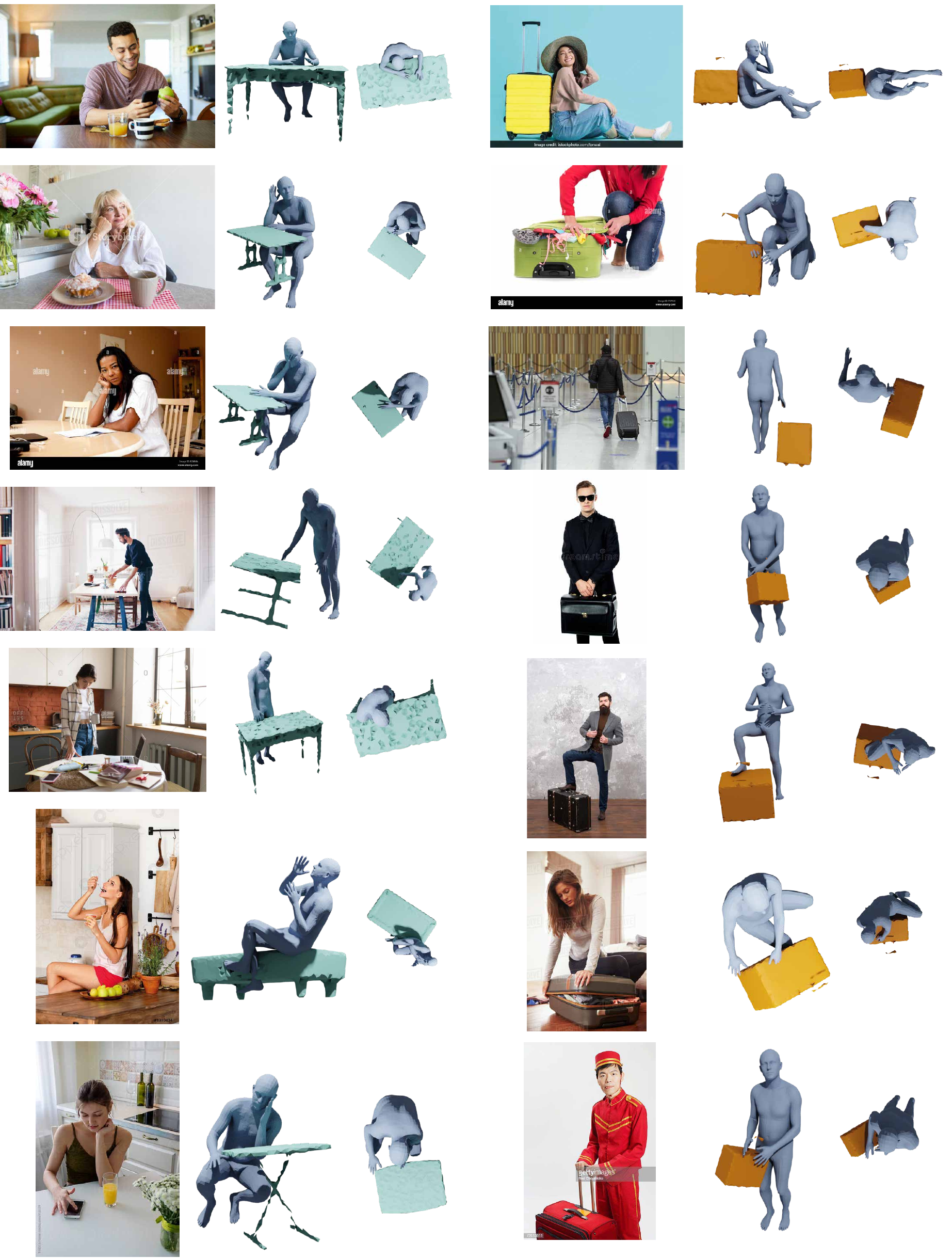}
    \caption{\textbf{Examples of inferred 3D interactions with table and suitcase.}}
    \label{fig:add1}
\end{figure*}

We provide additional results for our method in Figure~\ref{fig:add1} and Figure~\ref{fig:add2}. Figure~\ref{fig:add1} shows different interactions with table and suitcase and Figure~\ref{fig:add2} gives examples of interactions with other object categories in PartNet. 

\begin{figure*}
    \centering
    \includegraphics[width=\linewidth]{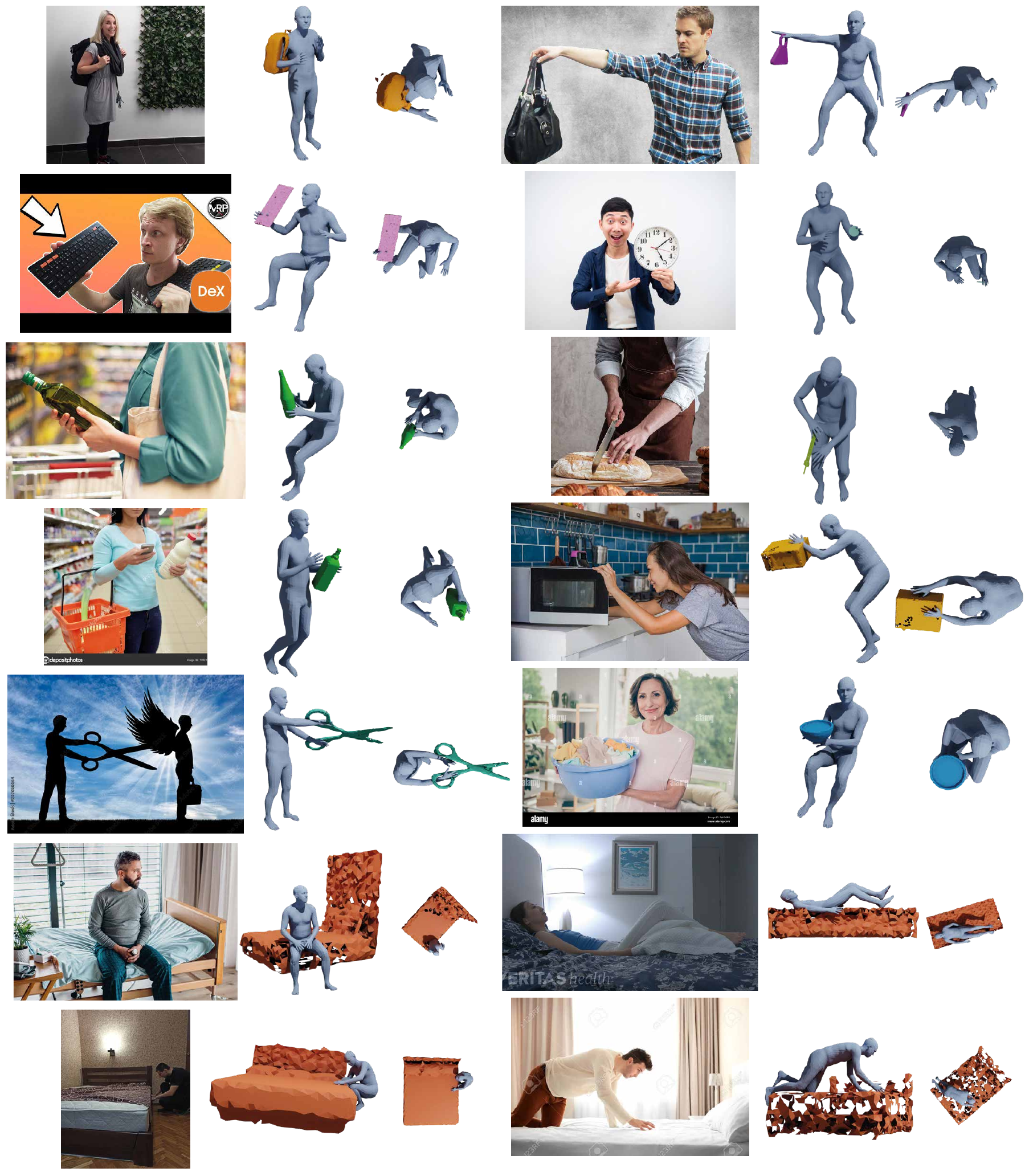}
    \caption{\textbf{Examples of inferred 3D interactions with several PartNet object categories.}}
    \label{fig:add2}
\end{figure*}

\end{document}